\documentclass{article}

\usepackage{PRIMEarxiv}

\usepackage[utf8]{inputenc} % allow utf-8 input
\usepackage[T1]{fontenc}    % use 8-bit T1 fonts
\usepackage{hyperref}       % hyperlinks
\usepackage{url}            % simple URL typesetting
\usepackage{booktabs}       % professional-quality tables
\usepackage{amsfonts}       % blackboard math symbols
\usepackage{nicefrac}       % compact symbols for 1/2, etc.
\usepackage{microtype}      % microtypography
\usepackage{lipsum}
\usepackage{fancyhdr}       % header
\usepackage{graphicx}       % graphics
\graphicspath{{media/}}     % organize your images and other figures under media/ folder
\usepackage{subfigure}
\usepackage{float}  %figure inside minipage
\usepackage{fancyhdr}
\usepackage{xcolor} % changed from color to xcolor (2021/11/24)
\usepackage{algorithm}
\usepackage{algorithmic}
\usepackage{natbib}
\usepackage{eso-pic} % used by \AddToShipoutPicture
\usepackage{forloop}
\usepackage{url}

\usepackage{amsmath}
\DeclareMathOperator*{\argmin}{argmin}

\usepackage{amssymb}
\usepackage{mathtools}
\usepackage{amsthm}

\usepackage[capitalize,noabbrev]{cleveref}
\setcitestyle{authoryear,round,citesep={;},aysep={,},yysep={;}}

\theoremstyle{plain}
\newtheorem{theorem}{Theorem}[section]

\newtheorem{lemma}[theorem]{Lemma}

\theoremstyle{definition}
\newtheorem{definition}[theorem]{Definition}

\theoremstyle{remark}

\title{Safe and Adaptive Decision-Making for Optimization of Safety-Critical Systems: The ARTEO Algorithm
}

\author{
  Buse Sibel Korkmaz \\
  Imperial College London \\
  London, United Kingdom\\
  \texttt{buse.korkmaz18@imperial.ac.uk} \\
   \And
  Marta Zagórowska \\
  ETH Zürich \\
  Zürich, Switzerland\\
  \texttt{mzagorowska@control.ee.ethz.ch} \\
  \And
  Mehmet Mercangöz \\
  Imperial College London \\
  London, United Kingdom\\
  \texttt{m.mercangoz@imperial.ac.uk} \\
}

\begin{document}
\maketitle

\begin{abstract}
We consider the problem of decision-making under uncertainty in an environment with safety constraints. Many business and industrial applications rely on real-time optimization to improve key performance indicators. In the case of unknown characteristics, real-time optimization becomes challenging, particularly because of the satisfaction of safety constraints. We propose the ARTEO algorithm, where we cast multi-armed bandits as a mathematical programming problem subject to safety constraints and learn the unknown characteristics through exploration while optimizing the targets. We quantify the uncertainty in unknown characteristics by using Gaussian processes and incorporate it into the cost function as a contribution which drives exploration. We adaptively control the size of this contribution in accordance with the requirements of the environment. We guarantee the safety of our algorithm with a high probability through confidence bounds constructed under the regularity assumptions of Gaussian processes. We demonstrate the safety and efficiency of our approach with two case studies: optimization of electric motor current and real-time bidding problems. We further evaluate the performance of ARTEO compared to a safe variant of upper confidence bound based algorithms. ARTEO achieves less cumulative regret with accurate and safe decisions.
\end{abstract}
% keywords can be removed
% \keywords{First keyword \and Second keyword \and More}

\section{INTRODUCTION}
In most sequential decision-making problems under uncertainty, there exists an unknown function with noisy feedback that the decision-maker algorithm needs to sequentially estimate and optimize to reveal the decisions leading to the highest reward. Each decision receives an instantaneous reward with an initially unknown distribution in a stochastic optimization setting. In this setting, initial decisions are made based on some heuristics, and incurred reward is memorized to exploit for the next decisions. Therefore, the uncertainty due to unknown characteristics decreases while new decisions are made based on previous reward observations. Even though more exploration can help to optimize the decisions by revealing more information in each iteration, it could be very expensive to evaluate the unknown function for many applications. Therefore, there is a need to balance exploring unknown decision points and exploiting previous experiences. The trade-off between exploration and exploitation is extensively studied in the literature, and the multi-armed bandit (MAB) approaches with confidence bounds are suggested to solve this problem by optimally balancing exploration and exploitation \citep{bubeck2012regret}.

\textbf{Related work.}\hspace{11pt}The idea of using confidence bounds to balance the exploration-exploitation trade-off through the optimism principle first appeared in the work of \citet{1985Lai} with the utilization of the upper confidence bound (UCB). Since then, it led to the development of UCB algorithms for stochastic bandits with many arms \citep{banditbook}. Many efficient algorithms build for bandit problems having a cost or reward function under certain regularity conditions \citep{Dani2008, Bubeck2008}. In \citet{Srinavas2010}, the authors divided the stochastic optimization problem into two objectives: (1) unknown function estimation from noisy observations and (2) optimization of the estimated function over the decision set. They used kernel methods and Gaussian processes (GPs) to model the reward function since these methods encode the regularity assumptions through kernels \citep{RasmussenGP}. Even though these methods are very successful to model and optimize unknown reward functions, they do not consider safety constraints.

In safety-critical systems, it is not possible to do exploration in some parts of the decision space due to safety concerns, which are modelled as safety constraints in optimization. Many industrial applications fall under the safety-critical systems due to their risk of  danger to human life, leading to substantial economic loss, or causing severe environmental damage. For example, we can consider a chemical process plant as a safety-critical operation since we need to satisfy the constraints of chemical reactions and surrounding processes to not cause a hazardous event for the environment and human operators. Therefore, if we want to approximate the unknown characteristics and optimize a process in this plant, we need to utilize safe exploration algorithms which allow only the exploration of feasible decision points by enforcing safety constraints. In this work, we define a feasible decision point in the given decision set as satisfying the safety and other constraints of the given problem.

The safe exploration problem has been studied by formalizing it as both bandit and Markov Decision Processes (MDPs) problems \citep{Schreiter2015, Turchetta2016, Wachi2018, Turchetta2019}. In \citet{Sui2015}, the authors have introduced the SafeOpt algorithm and showed it is possible that safely optimize a function with an unknown functional form by creating and expanding a safe decision set through safe exploration under certain assumptions. They also provided the safety guarantee of SafeOpt by using the confidence bounds construction method of \citet{Srinavas2010}. These safe exploration algorithms are applied to many control and reinforcement learning problems and proved their success in those domains \citep{app1,app2}. However, these algorithms either require an exploration phase or apply a trade-off strategy between optimal decisions and exploration. There are some real-world applications such as industrial processes which cannot afford an exclusive exploration phase or require following optimization goals of the system even in the explored points. For example, any deviation from target satisfaction in an industrial process may cause high costs to the plant or damage the reputation of the responsible party. Thus, there is a need to consider the exploration in an adaptive manner in accordance with the requirements of the environment and this aspect is not covered by existing safe exploration algorithms.

\textbf{Our contributions.}\hspace{11pt}In this paper, we propose a novel safe Adaptive Real-Time Exploration and Optimization (ARTEO) algorithm, where we cast multi-armed bandits as a mathematical programming problem subject to safety constraints for the optimization of safety-critical systems. Our contribution is posing a further constraint to the safe exploration by instructing the algorithm to avoid decisions not satisfying the optimization goals of the safety application. This constraint discourages exploration which makes it harder to learn unknown rewards in the MAB setting. To compensate for that, we incorporate the exploration as a contribution to the cost function in an adaptive manner by encouraging explicitly to take decisions at points with high uncertainty which is quantified by the covariance of GP models. We adaptively control the size of this contribution in accordance with the requirements of the environment. ARTEO uses GPs to model how the unknown parameters of an application change based on external factors such as inputs and optimization goals, make new decisions when a change in the optimization goals or in the given input is occurred and explores new decision points with either a change in the optimization goals or based on uncertainty. We establish the safety of ARTEO by constructing the confidence bounds as \citet{Srinavas2010} and showing certain assumptions hold in our setting. We demonstrate the safety and efficiency of our algorithm using two real-world examples: the electric current optimization in electric motors and the high-dimensional online bid optimization problem. In the first experiment, the results show that we are able to successfully estimate and optimize the current without any constraint violation by applying ARTEO. In the second experiment, our approach satisfies the constraints while yielding more profitable outcomes.

\section{PROBLEM STATEMENT AND BACKGROUND}

% We want to find a sequence of decisions, $x_1,x_2, \ldots, x_T$, such that a certain cost function $f$ is minimised. At every iteration $t$, $t=1,\ldots,T$, the cost function depends on the decision $x_t \in D$, and on the system characteristics $v_t\in P$ where $P \subset \mathbb{R}^u$ and $u$ denotes the number of unknowns. $D \subset \mathbb{R}^n$ is a decision set for the decision variable $x$ and $n$ represents the number of decision variables. The characteristics $v_t$ depends on the decision $x_t$, i.e. $v_t=p(x_t)$ where $p:D \rightarrow  P$. After making a decision $x_t$, we obtain a noisy measurement, $y_t=f(x_t,p(x_t))+\epsilon$ where $f:D \times P \rightarrow \mathbb{R}$ and $\epsilon$ is $R$-sub-Gaussian noise with a fixed constant $R \geq 0$ \citep{SubGaussian}. We assume that we know the functional form of $f(\cdot)$, but the functional form of $p(\cdot)$ is unknown. Furthermore, at every iteration, the decision $x_t$ must satisfy the  constraints $g_a(x_t,p(x_t))+ h_a\leq 0$, where $a=1,2,\ldots, A$ with $A$ denoting the number of constraints. The value of $h_a\geq 0$ is called a safety threshold. Thus, we can formalise our optimization problem at time $t$ as:
We want to find a sequence of decisions, $x_1, x_2, \ldots, x_T$, so that a certain cost function $f$ is minimised. At each iteration $t$, where $t = 1, \ldots, T$, the cost function is dependent on the decision $x_t \in D$, and on the system characteristics $v_t \in P$. Here, $P \subset \mathbb{R}^d$ and $\Lambda$ is the set of unknowns $\Lambda=\left\{\lambda=1,\ldots,d\right\}$ with $d$ elements, where $\lambda$ is the index of an unknown, and $D \subset \mathbb{R}^n$, with $n$ being the number of decision variables. The characteristics $v^\lambda_t$ are determined by the decision $x_t$, i.e. $v^\lambda_t = p_\lambda(x_t)$ where $p:D \rightarrow P$. After making a decision $x_t$, we obtain a noisy measurement, $y_t = f(x_t,p_\Lambda(x_t)) + \epsilon$, where $f:D \times P \rightarrow \mathbb{R}$ and $\epsilon$ is an $R$-sub-Gaussian noise with a fixed constant $R \geq 0$ \citep{SubGaussian}. We assume that we know the functional form of $f(\cdot)$, but the functional form of $p_\lambda(\cdot)$ is unknown. Furthermore, at every iteration, the decision $x_t$ must satisfy the  constraints $g_a(x_t,p_\Lambda(x_t)) \leq h_a$, where $a=1,2,\ldots,A$ with $A$ denoting the number of constraints. The value of $h_a$ is called a safety threshold. Thus, we can formalise our optimization problem at time $t$ as:
\begin{equation}\label{opt_problem}
\begin{aligned}[c]
 X^*_t=&{\argmin_{X} f(X_t,p_\Lambda(X_t)) \text{ s.t. } g_a(X_t,p_\Lambda(X_t)) \leq h_a,  \forall a}
\end{aligned}
\end{equation}
where $X = [x_{i}, ..., x_{n}]$ are decision variables. If we know $p_\Lambda(x_t)$, we can solve the problem as an optimization problem with noise. For instance, we could use the concept of real-time optimization (RTO) from the process control domain \citep{RTOChemical} and use the approach proposed by \citet{SafeRTO}. However, the characteristics $p_\Lambda(\cdot)$ are unknown. Thus, at every iteration $t$, we first solve an estimation problem to find $p$, then solve the optimization problem (\ref{opt_problem}). Solving an optimization problem by combining estimation then optimization is a common approach \citep{EstimationforRTO1, EstimationforRTO2}. However, few approaches quantify the uncertainty inherent in the estimation of $p$. In the current paper, we propose to estimate $p$ using Gaussian processes and use regularity assumptions of Gaussian processes to ensure safety.

\subsection{Gaussian Processes} 

Gaussian processes are non-parametric models which can be used for regression. GPs are fully specified by a mean function $\mu(x)$ and a kernel  $k(x,x')$ which is a covariance function and specifies the prior and posterior in GP regression \citep{RasmussenGP}. The goal is to predict the value of the unknown characteristics $p$ over the decision set $D$ by using GPs to solve the optimization problem in (\ref{opt_problem}). Assuming having a zero mean prior, the posterior over $p$ follows $\mathcal{N}(\mu_T(x), \sigma^2_T(x))$ that satisfy, 
\begin{equation}\label{GPposterior}
\begin{aligned}[c]
  \mu_T(x) = &{k_T(x)^T(K_T + \sigma^2I)^{-1} y_T}\\
  k(x, x') = &{k (x, x') - k_T(x)^T(K_T+\sigma^2I)^{-1}k_T(x')} \\ 
  \sigma^2(x) = &{k_T(x, x)}
\end{aligned}
\end{equation}

where $k_T(x) = [k(x_1, x), \ldots , k(x_T, x)]$, and $K_{T}$ is the positive definite kernel matrix $[k(x, x')]_{x,x' \in \left\{x_1, \ldots, x_T\right\}}$. By using GPs, we can define estimated $\hat{p_\lambda}$ at $x_t$ with a mean $\mu_{\hat{p_\lambda}}(x_t)$ and standard deviation $\sigma_{\hat{p_\lambda}}(x_t)$.

\subsection{Regularity Assumptions}\label{sec:2.2}

We do not have any prior knowledge of how $f$ and $g_a$ change based on external factors such as the optimization goals or inputs, and to provide safety with high probability at decision points we need to make some assumptions \citep{Srinavas2010, Sui2015, app1,Sui2018}. For simplicity, we continue as such we have one safety constraint ($A=1$) and one unknown ($d=1$). We represent them as $g$ and $\hat{p}$ without index. However, all assumptions we have in this section can be applied to any safety function $g_a$ and unknown $\hat{p}_\lambda$ when the optimization problem consists of multiple safety constraints and unknowns. 

We assume the decision set $D$ is compact as being a closed and bounded subset of Euclidean space \citep{Compactness}. Furthermore, the cost function $f$ and safety function $g$ might include known terms $\Delta$, which are assumed to be continuous over $D$, besides unknown characteristics $p$. We sample unknown characteristics $\hat{p}$ from a GP with a positive definite kernel, which means $\hat{p}$ is continuous by definition of positive definite kernels \citep{RasmussenGP}. Next, we introduce a lemma for the continuity of $f$ and $g$.

\begin{lemma} \label{lem:lemma1} [adapted from \citet{continuityoverfunctions}] Let $f$ be a function of known terms $\Delta(\cdot)$ and unknown terms $p(\cdot)$. $p$ and $\Delta$ are continuously defined in domain $D$. Given $p$ and $\Delta$ are continuous in $D$, any $f$ function that is formed by an algebraic operation over two functions $p$ and $\Delta$ is also continuous in $D$. 
\end{lemma}

Following \cref{lem:lemma1}, the continuity assumption holds for cost function $f$ and safety function $g$ since they are formed by continuous terms. Next, we define the relationship between $g$ and $\hat{p}$.

\begin{definition}\label{def:monotonicallyrelated} \textit{(monotonically related)} A function $\phi(\cdot, \pi(x))$ is monotonically related to $\pi(x)$ if $\phi$ and $\pi$ are continuous in $D$ and for any $x,y \in D$ such that $\pi(x) \leq \pi(y) \Rightarrow \phi(\cdot, \pi(x))\leq \phi(\cdot, \pi(y))$.
\end{definition}

\begin{definition}\label{def:inverselymonotonicallyrelated} \textit{(inversely monotonically related)} A function $\phi(\cdot, \pi(x))$ is inversely monotonically related to $\pi(x)$ if $\phi$ and $\pi$ are continuous in $D$ and for any $x,y \in D$ such that $\pi(x) \leq \pi(y) \Rightarrow \phi(\cdot, \pi(x)) \geq \phi(\cdot, \pi(y))$.
\end{definition}

We assume $g$ is monotonically related or inversely monotonically related to $\hat{p}$ as in \cref{def:monotonicallyrelated} and \cref{def:inverselymonotonicallyrelated}. This assumption allows us to reflect confidence bounds of $\hat{p}$ to $g$. The continuity assumptions and ability to provide confidence bounds for $p$ depend on which model is used to estimate $p$. In this paper, we choose GPs which are related to reproducing kernel Hilbert space (RKHS) notion through their positive semidefinite kernel functions \citep{kernelcharacteristics} that allow us to construct confidence bounds in a safe manner later. 

The RKHS which is denoted by $\mathcal{H}_k(D)$ is formed by ``nice functions'' in a complete subspace of $L_2(D)$ and the inner product $\langle\cdot,\cdot\rangle_k$ of functions in RKHS follows the reproducing property: $\langle p,k(x,\cdot)\rangle_k = p(x)$ for all $p \in \mathcal{H}_k(D)$. The smoothness of a function in RKHS with respect to kernel function $k$ is measured by its RKHS norm $\|p\|_k = \sqrt{\langle p,p\rangle_k}$ and for all functions in $\mathcal{H}_k(D)$ $\|p\|_k < \infty$ \citep{Scholkopf2002}. Thus, we assume a known bound $B$ for the RKHS norm of the unknown function $p$: $\|p\|_k < B$. We use this bound $B$ to control the confidence interval (CI) width later in \cref{betateq}. In most cases, we are not able to compute the exact RKHS norm of the unknown function $p$ as stated by previous studies \citep{jiao2022backstepping}. Alternative approaches are choosing a very large $B$, or obtaining an estimate for $B$ by guess-and-doubling. It is possible to apply hyperparameter optimization methods to optimize $B$ where data is available offline \citep{berkenkamp2019no}. Since ARTEO utilizes an online learning concept, we choose a large $B$ in our case studies.

\subsection{Confidence Bounds} 

In ARTEO, we give safety constraints to the mathematical programming solver as hard constraints. The solver uses the CI which is constructed by using the standard deviation of conditioned GP on previous observations to decide the feasibility of a chosen point $x_t$. Hence, the correct classification of decision points in $D$ relies on the confidence-bound choice. Under the regularity assumptions stated in \cref{sec:2.2}, Theorem 3 of \citet{Srinavas2010} and Theorem 2 of \citet{BoundProof} proved that it is possible to construct confidence bounds which include the true function with probability at least $1-\delta$ where $\delta \in (0,1)$ on a kernelized multi-armed bandit problem setting with no constraints. Moreover, as shown by \citet{Sui2018} in Theorem 1, this theorem is applicable to multi-armed bandit problems with safety constraints. Hence, we can state that the probability of the true value of $p$ at the decision point $x_t$ is included inside the confidence bounds in iteration $t$:
\begin{equation}\label{confidenceboundsofprobabilityp}
P\bigr[|p(x_t) - \mu_{t-1}(x_t)| \leq  \beta_t\sigma_{t-1}(x_t)\bigr] \leq 1 - \delta, \ \forall t \geq 1
\end{equation}
where $\mu_{t-1}(x_t)$ and $\sigma_{t-1}(x_t)$ denote the mean and the standard deviation at $x_t$ from a GP at iteration $t$, which is conditioned on previous $t-1$ observations to obtain the posterior. $\delta$ is a parameter that represents the failure probability in \cref{confidenceboundsofprobabilityp}. $\beta_t$ controls the width of the CI and satisfies \cref{confidenceboundsofprobabilityp} when:
\begin{equation}\label{betateq}
\beta_t = B + R\sqrt{2(\gamma_{t-1} + 1 + \text{ln}(1/\delta))}
\end{equation}
where the noise in observations is $R$-sub-Gaussian and $\gamma_{t-1}$ represents the maximum mutual information after $t-1$ iterations. $\gamma_{t}$  is formulated as:
\begin{equation}
\gamma_{t} = \max_{|X_{t}| \leq t} I(\hat{p}; y_{t})
\end{equation}
where $y_t$ represents the evaluations of $\hat{p(\cdot)}$ at decision points $x = x_{1 \ldots t}$. $I(\cdot)$ denotes the mutual information as such:
\begin{equation}
I(p; y_t) = 0.5 \log|\text{I} + \sigma^{-2}K_{t}|
\end{equation}

\cref{confidenceboundsofprobabilityp} gives us the confidence bounds of $\hat{p}$. However, in order to establish safety with a certain probability, we need to obtain confidence bounds of $g(x_t)$ for each iteration. To expand \cref{confidenceboundsofprobabilityp} for $g$, we provide the following lemma.

\begin{lemma} \label{lem:lemma3} Let $\hat{p}^L = \mu(x_t) - \beta_t \sigma_{t}(x_t)$ and $\hat{p}^U = \mu(x_t) + \beta_t \sigma_{t}(x_t)$ where $\hat{p}^L \leq \hat{p}(x) \leq \hat{p}^U$. Given $g$ is monotonically related to $\hat{p}$, $g$ is in a known functional form of $g(\Delta(x), \hat{p}(x))$ with a known value of $\Delta(x)$, and $\hat{p}^L \leq \hat{p}(x) \leq \hat{p}^U$:
\begin{itemize}
    \item if $g$ is monotonically related to $\hat{p}$ $\Rightarrow$ $g(\Delta(x), \hat{p}^L) \leq g(\Delta(x), \hat{p}(x)) \leq g(\Delta(x), \hat{p}^U)$.
    \item if $g$ is inversely monotonically related to $\hat{p}$ $\Rightarrow$ $g(\Delta(x), \hat{p}^U) \leq g(\Delta(x), \hat{p}(x)) \leq g(\Delta(x), \hat{p}^L)$.
\end{itemize}
\end{lemma}

Now, we present the main theorem that establishes the safety of ARTEO with high probability based on regularity assumptions and \cref{lem:lemma3}. The proofs for \cref{lem:lemma3} and \cref{thm:safetytheorem} is given in \cref{appendixproofs}.

\begin{theorem}\label{thm:safetytheorem} Suppose that $\hat{p}$ and $g$ are continuous on compact set $D$, the functional form of $g(\Delta(x), \hat{p}(x))$ is known and $g$ is monotonically related to $\hat{p}$ where $\hat{p}$ is modelled from a GP through noisy observations $y_t = \hat{p}(x_t) + \epsilon_t $ and $\epsilon_t$ is a $R$-sub-Gaussian noise for a constant $R \geq 0$ at each iteration $t$. For a known value of $\Delta(x)$, the maximum and minimum values of $g(\Delta(x), \hat{p}(x))$ lie on the upper and lower confidence bounds of the Gaussian process obtained for $\hat{p}(x)$ which are computed as in \cref{lem:lemma3}. For a chosen $\beta_t$ and allowed failure probability $\delta$ as in \cref{betateq}, 
\begin{itemize}
    \item if $g$ is monotonically related to $\hat{p} \Rightarrow P\bigr[ g(\Delta(x_t), \hat{p}^L) \leq g (\Delta(x_t), \hat{p}(x_t)) \leq g(\Delta(x_t), \hat{p}^U) \bigr] \leq 1 - \delta, \forall t \geq 1$
    \item if $g$ is inversely monotonically related to $\hat{p} \Rightarrow P\bigr[ g(\Delta(x_t), \hat{p}^U) \leq g (\Delta(x_t), \hat{p}(x_t)) \leq g(\Delta(x_t), \hat{p}^L) \bigr] \leq 1 - \delta, \forall t \geq 1$
\end{itemize}
\end{theorem}

\section{ARTEO ALGORITHM}

We develop the ARTEO algorithm for safety-critical environments with high exploration costs. At each iteration, the algorithm updates the posterior distributions of GPs with previous noisy observations as in \cref{GPposterior} and provides an optimized solution for the desired outcome based on how GPs model the unknown components. It does not require a separate training phase, instead, it learns during normal operation. The details of safe learning and optimization are given next. 

\subsection{Safe Learning}

The decision set $D_i$ is defined for each variable $i$ as satisfying the introduced assumptions in Section 2. For each decision variable, a GP prior and initial ``safe seed'' set is introduced to the algorithm. The safe seed set $S_0$ includes at least one safe decision point with the true value of the safety function at that point satisfying the safety constraint(s). As in many published safe learning algorithms \citep{Sui2015, Sui2018, Turchetta2019}, without a safe seed set, an accurate assessment of the feasibility of any points is difficult. Each iteration of the algorithm could be triggered by time or an event. After receiving the trigger, the algorithm utilizes the past noisy observations to obtain the GP posterior of each unknown to use in the optimization of the cost function, which includes the cost of decision and uncertainty. For the first iteration, safe seed sets are given as past observations. 

The uncertainty in the cost function $f$ is quantified as:
\begin{equation}\label{uncertaintycalc}
\begin{aligned}[c]
    U(X_t, \hat{p}_\Lambda(X_t)) = &{\sum_{\lambda=1}^{d}{\sigma_{\hat{p}_\lambda}(X_{t})}}
\end{aligned}
\end{equation}
where $\sigma_{\hat{p}_\lambda}(X_{t})$ is the standard deviation of $GP_\lambda$ at $X_{t}$ for the unknown $\lambda$ in the iteration $t$. It is incorporated into the $f$ by multiplying by an adjustable parameter $z$ as next:
\begin{equation}
    f(X_t, \hat{p}_\Lambda(X_t)) = C(X_t, \hat{p}_\Lambda(X_t)) - z U(X_t, \hat{p}_\Lambda(X_t))
\end{equation}
where $C(X_t, \hat{p}_\Lambda(X_t))$ represents the cost of decision at the evaluated points. In our experiments, the priority of the algorithm is optimizing the cost of decisions under given constraints. Hence, the uncertainty weight $z$ remains zero until the environment becomes available for exploration. Until that time, the algorithm follows optimization goals and learns through changes in the optimization goals such as operating in a different decision point to satisfy a new current in our first case study.

\begin{figure}[t]
\begin{algorithm}[H]
\caption{ARTEO}\label{alg:ARTEO}
\begin{algorithmic}
 \STATE {\bfseries Input:} Decision set $D_i$ for each variable $i \in \left\{1,..,n\right\}$, GP priors for each ${GP}_\lambda$, safe seed set for each GP as $ S_{\lambda,0}$, uncertainty weight as $\zeta$, cost function $f$, safety function $g$, safety threshold $h$, the exploration heuristic rule
\FOR{$t=1,...,T$} 
\IF{ the exploration heuristic rule holds}
\STATE $z \gets \zeta$
\ELSE{}
\STATE $z \gets 0$ 
\ENDIF
\FOR{$\lambda=1,..., d$}
\STATE Update $\hat{p}_\lambda$ by conditioning ${GP}_\lambda$ on $S_{\lambda,t-1}$ 
\ENDFOR
\STATE $x_{\left\{1, ..., n\right\},t}^* \gets \operatorname*{argmin}_{x_i \in D_i}{f \text{ s.t. } g }$ 
\FOR{$\lambda=1,..., d$}
\STATE $y_{\lambda,t} \gets$ $\hat{p}_{\lambda}(X_{t}^*) + \epsilon_t$
\STATE $S_{\lambda,t} \gets $ $S_{\lambda,t-1} \cup \left\{X_t^*:{y_{\lambda,t}}\right\} $ 
\ENDFOR
\ENDFOR
\end{algorithmic}
\end{algorithm}
\end{figure}

The exploration is controlled by the $z$ hyperparameter. It is possible to create different heuristic rules for changing the $z$ parameter during the execution to fine-tune it based on the needs of the simulated environment. This change in the cost function encourages the real-time optimizer to take decisions on unexplored points to decrease uncertainty as a part of the optimization. The safety constraints and the aim of satisfying the expectation with a minimum cost are still under consideration by the RTO during this phase, so, the algorithm does not violate safety constraints with a $1 - \delta$ probability as explained in the previous section. The pseudocode of the algorithm is given in \cref{alg:ARTEO}.

\subsection{Optimization}

The RTO incorporates the posterior of the Gaussian process into decision-making by modelling the unknown characteristics by using the mean and standard deviation of GPs. The cost function is the objective function in the RTO formulation and the safety thresholds are constraints. In the cost function, the mean of GP posterior of each uknown is used to evaluate the cost of decision and the standard deviation of GPs is used to measure uncertainty as in \cref{uncertaintycalc}. In the safety function, the standard deviation of the GP posterior of each unknown is used to construct confidence bounds and then these bounds are used to assess the feasibility of evaluated points. The optimizer solves the minimization problem under safety constraints within the defined decision set of each decision variable. Any optimization algorithm that could solve the given problem can be used in this phase. 

\subsection{Complexity}

In each iteration, ARTEO updates GP models by conditioning GPs on past observations and finds a feasible solution. The overall time complexity of each run of the algorithm is the number of iterations $t$ times the time complexity of each iteration. The first computationally demanding step in the ARTEO is fitting GPs on safe sets. The time complexity of training a full GP, i.e. exact inference, is $\mathcal{O}((t-1)^3)$ due to the matrix inversion where $t-1$ is the number of past observations at iteration $t$. \citep{GPcovariancecomplexity}. It is possible to reduce it further by using low-rank approximations which is not in the scope of our work \citep{GPlowrank1, GPlowrank2}. We introduce an individual GP for each unknown, so the total complexity of GP calculations is $\mathcal{O}(d(t-1)^3)$.

The next demanding step is nonlinear optimization. The computational demand of RTO depends on the chosen optimization algorithm and the required $s$ number of steps to converge. The most computationally expensive step  is the LDL factorization of a matrix with a $\mathcal{O}((t-1)^3)$ complexity in both used optimization algorithms in this paper \citep{SLSQP,interiorpointmethod}. Hence, the complexity of RTO becomes $\mathcal{O}(s(t-1)^3)$. In our implementation, $s \gg d$, so, the time complexity of each iteration in ARTEO scales with the RTO complexity. Therefore, the overall time complexity of one iteration of ARTEO is $\mathcal{O}(s(t-1)^3) \approx \mathcal{O}(st^3)$. The memory complexity of the algorithm is $\mathcal{O}(d(t-1)^2) \approx \mathcal{O}(dt^2)$, which is dominated by matrix storage in GPs and optimization.

\subsection{Limitations}

ARTEO shares a common limitation amongst algorithms using constraint-based solvers: the initialization problem for starting the optimization \citep{Initialization}. To address this, we use safe seeds as a feasible starting point for the optimization routine and leverage $x_{t-1}^*$ as an initial guess for subsequent iterations. However, if $x_{t-1}^*$ violates the safety constraints (may happen with a probability of $\delta$), the success of the optimization at time $t$ depends on whether the chosen solver can take an infeasible guess as a starting point. Even though many solvers can handle infeasible starting points, they are mostly local solvers which means that if we start far from the actual solution we may reach only a local minimum and it may take a significant amount of time to converge to a solution. Additionally, the time complexity of ARTEO may cause it to be unsuitable for environments that require faster results.

\section{EXPERIMENTS}

In this section, we evaluate our approach on two applications: an electric motor current optimization and online bid optimization. The former problem is introduced in \citet{GEM} and the latter one in \citet{ipinyoudataset}. We develop the first case study with Matlab and the second one with Python on an M1 Pro chip with 16 GB memory. The GitHub link is available at \url{https://github.com/buseskorkmaz/ARTEO}.

\subsection{Electric Motor Current Optimization}

In this case study, we implement ARTEO to learn the relationship between torque and current in Permanently-Excited Direct Current Motors (PEDCMs), which have a positively correlated torque-current relationship. We develop the simulation with two PEDCMs as explained in \cref{gem-details} in Gym Electric Motor (GEM) \citep{GEM}. Then, the environment is simulated, and collected sample data points of torque and current are served as ground truth in our algorithm.

\subsubsection{ARTEO Implementation to Electric Motor Current Optimization}

We put the electric motor current optimization into our framework as following a reference current signal for alternator mode operation where the produced current at a given torque is initially unknown to ARTEO. The objective function is defined as 
\begin{equation}
    f(X_t) = \bigl[Cr_t - \sum_{\lambda=1}^{2}{\mu_{TC_\lambda}(X_{t})}\bigr]^2 - z \sum_{\lambda=1}^{2} {\sigma_{TC_\lambda}(X_{t})}
\end{equation}
where $X_{t} = [x_{1,t}, x_{2,t}]$ is the optimized torque of the electric motors for a given reference current $Cr_t$ at time $t$. $\mu_{TC_\lambda}(X_t)$ and $\sigma_{TC_\lambda}(X_t)$ represent the mean and standard deviation of GP regression for the unknown $\hat{p}_{TC_\lambda}$ of produced electric current for machine $\lambda$ at torque $X_{t}$. $z$ is the hyperparameter for driving exploration. The operation range limit of torque is implemented as bound constraints
\begin{equation}
     0 \leq x_{t\lambda} \leq 38 \text{ Nm} \quad \forall t,\lambda
\end{equation}
Lastly, the safety limit of the produced current for chosen electric motors is decided as 225.6 A according to the default value in the GEM environment and $g$ is formulated as
\begin{equation}
    \sum_{i=1}^{2}{\mu_{TC_i}(x_{ti}) + \beta_t \sigma_{TC_i}(x_{ti}) } \leq 225.6 \text{ A}
\end{equation}
where $\beta_t$ decides the width of the CI and is chosen as a value that satisfies \cref{betateq}. 
\begin{figure}[t]
  \centering
  \includegraphics[width=0.45\textwidth]{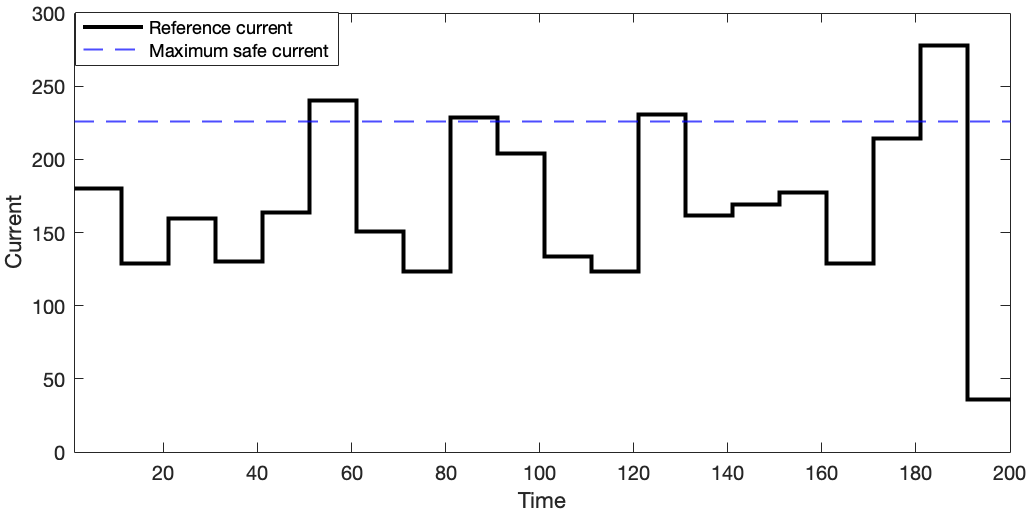}
  \caption{Reference current to distribute over two electric motors.}
  \label{Figure1}
\end{figure}

After building the optimization problem, the reference current to follow by ARTEO is generated as in \cref{Figure1}. The reference trajectory is designed in such a way that it includes values impossible to reach without violating the safety limit (the black points over the blue dashed line). The aim of this case study is to follow the given reference currents by assigning the torques to the motors while the current to be produced at the decided torque is predicted by the GPs of each motor. Therefore, ARTEO learns the torque-current relationship first from the given safe seed for each motor, then update the GPs of the corresponding
motors with its noisy observations at each time step.

\begin{figure}[b]
  \centering
  \includegraphics[width=0.45\textwidth]{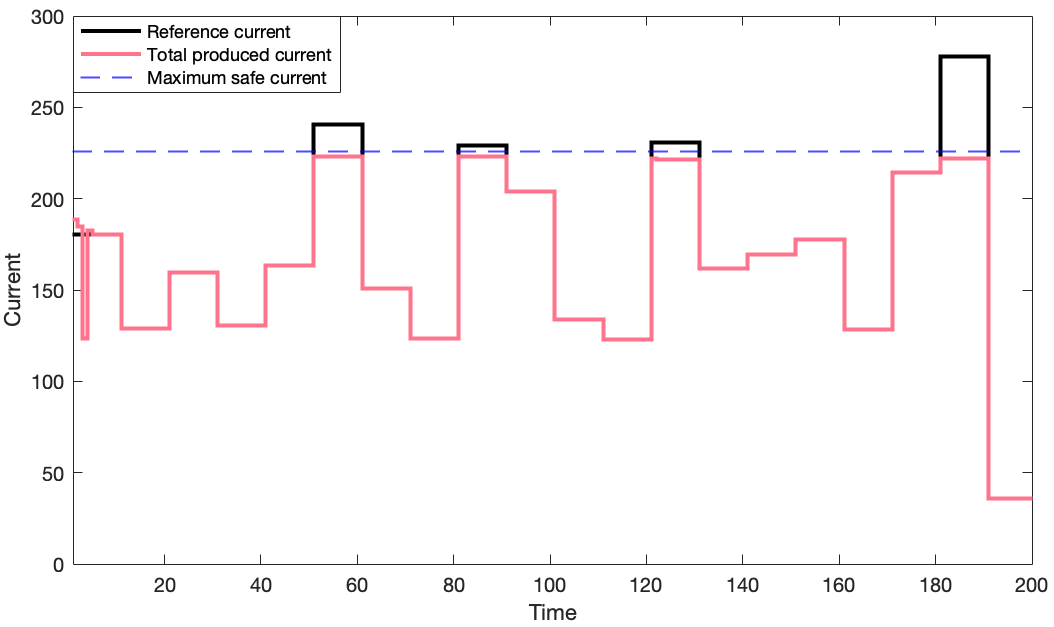}
  \caption{The results of ARTEO for the given reference current.}
  \label{Figure2}
\end{figure}

\begin{figure*}[t]
\begin{minipage}{\textwidth}
\begin{figure}[H]
    % \centering
    %  \vspace{-0.4cm}
    \includegraphics[width=8.4cm]{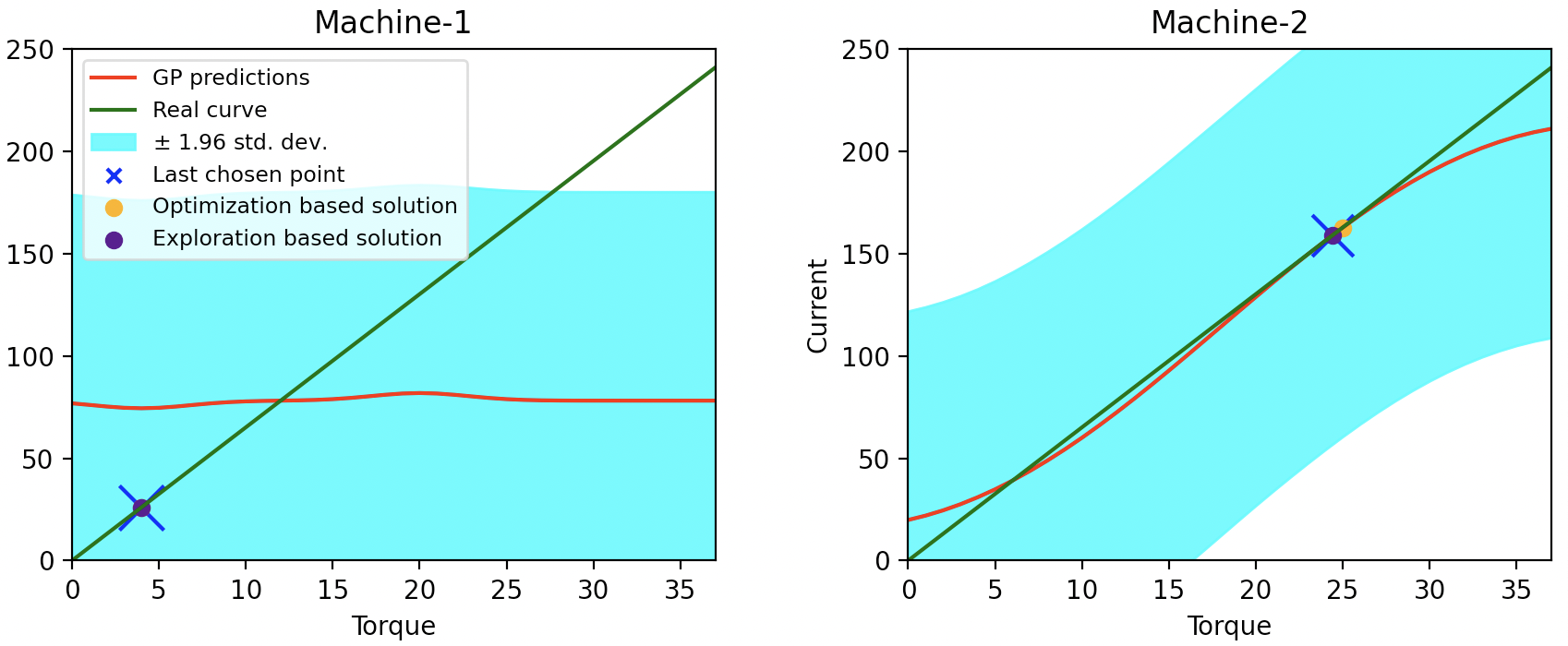}\label{a}
    \includegraphics[width=8.4cm]{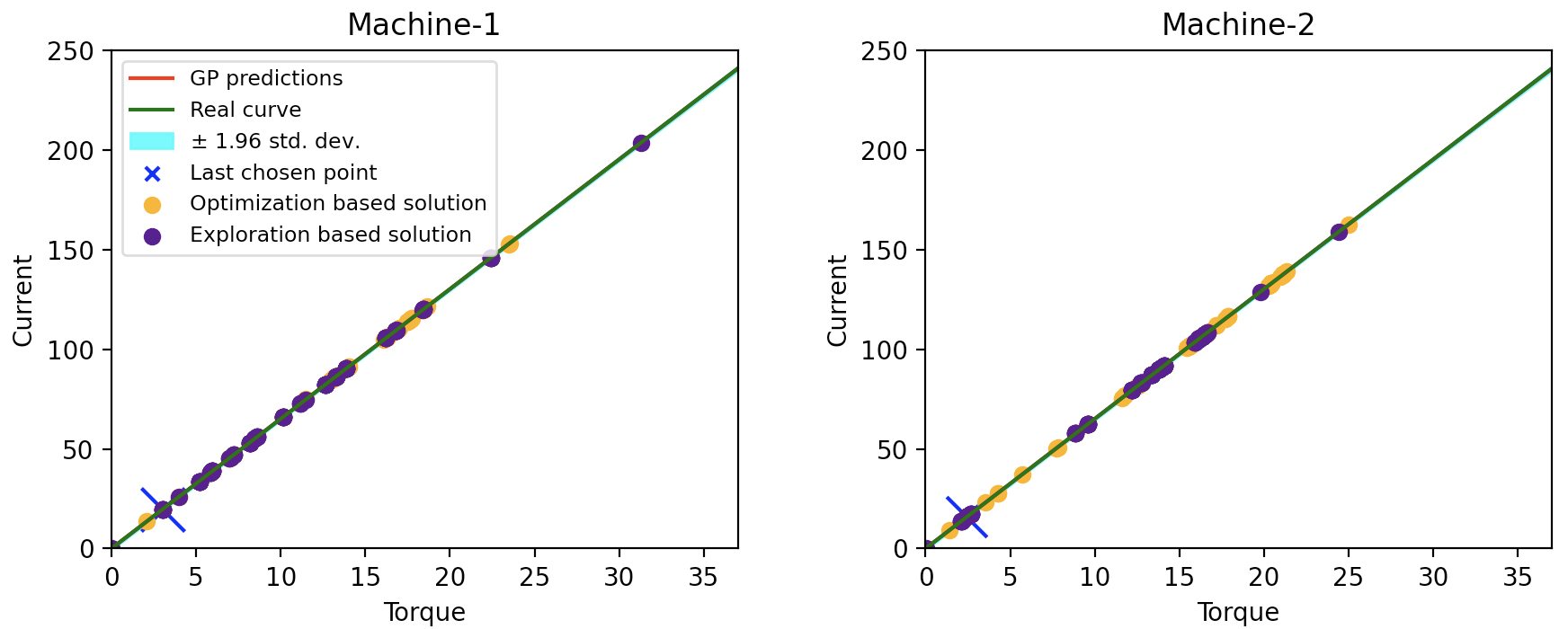}\label{b}
    % \vspace{-0.2cm}
    \caption{The second and last time steps of the simulation. The safe seed set includes two points for each electric motor within the operating interval. The blue-shaded area represents the uncertainty, which is high at the beginning due to unknown regions. GP-predicted torque-current lines are converged to the actual curves of each electric motor. Purple-coloured sample points are chosen by exploration.}
    \label{Figure3}
\end{figure}
\end{minipage}
\end{figure*}

The safe seed of each motor consists of two safe points which are chosen from collected data in the GEM environment. The kernel functions of both GPs are chosen as a squared-exponential kernel with a length scale of 215 (set experimentally). The value of exploration parameter $z$ is decided as 25. Hyperparameter optimization techniques for $z$ are discussed and employed in this case study in \cref{hyperparameteropt}. The result of the simulation for the reference in \cref{Figure1} is given \cref{Figure2}. \cref{Figure2} shows that ARTEO is able to learn the torque-current relationship for given electric motors and optimize the torque values to produce given reference currents after a few time steps without violating the maximum safe current limit for this scenario. 

The environment is set to be available for exploration if the reference current of time step $t$ is the same as the previous step and this reference is satisfied with a small margin ($\pm$ 5 A) in the previous step. When exploration starts (second time step) ARTEO prioritizes safe learning for the given $z$ value. In the second time step, ARTEO sends Machine-2 to a greater torque which helps decrease uncertainty while sending Machine-1 to a smaller torque to not violate the maximum safe current threshold. The comparison of estimated and real torque-current curves for the second and last time steps of the simulation is given in \cref{Figure3}. The effect of the exploration hyperparameter and recommended approaches to set it to an optimal value is discussed with additional experiments in \cref{detailsCurrentOpt}.

\subsubsection{Comparison with UCB algorithms}

\begin{figure}[tbhp]
  \centering
  \includegraphics[width=0.45\textwidth]{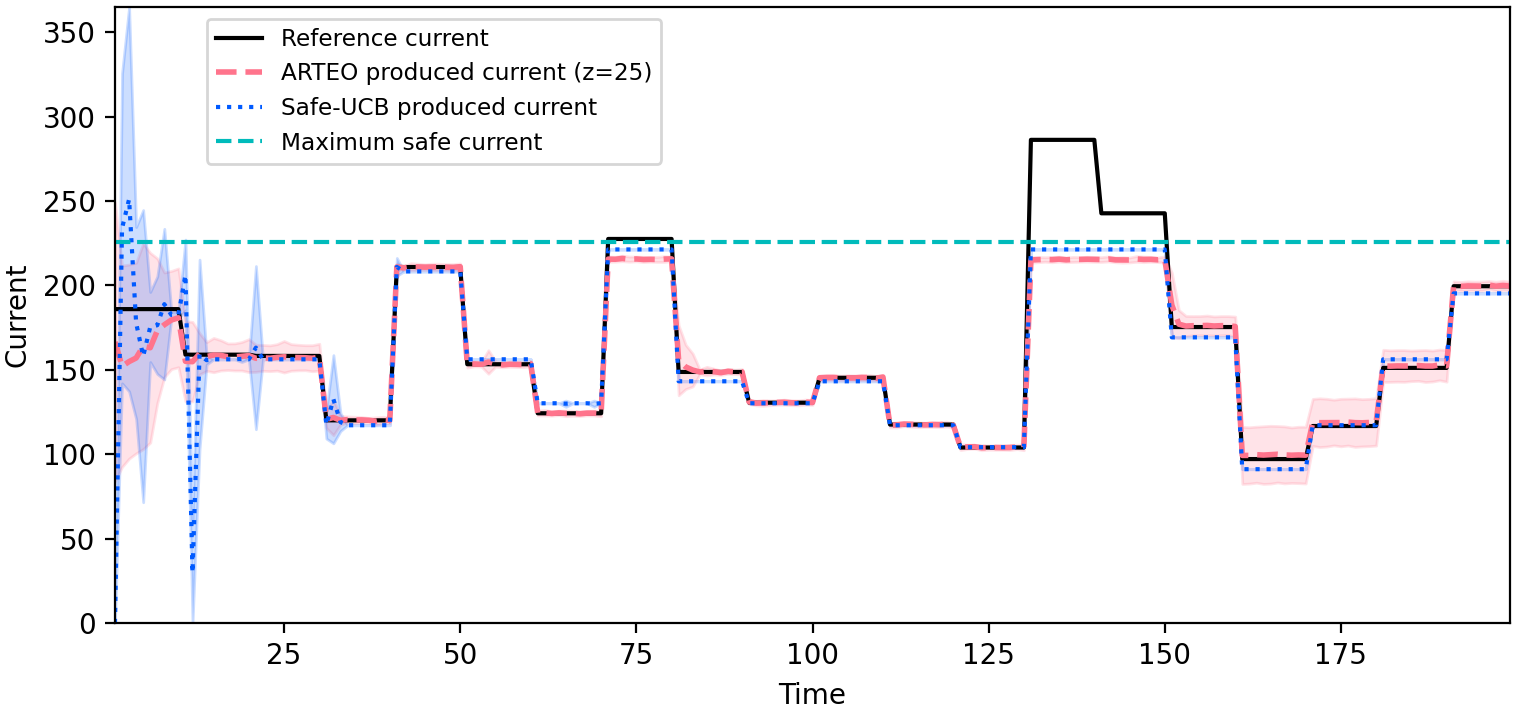}
  \caption{Comparison of produced currents of ARTEO and Safe-UCB algorithms. Shaded areas demonstrate $\pm1$ standard deviation added version of the same-colour used algorithm.}
  \label{Figure6}
\end{figure}

We compare ARTEO with a safety-aware version of GP-UCB algorithm. We implement GP-UCB as in \citet{Srinavas2010} to explore and exploit the decisions that minimize the objective function. Since the functional form is unknown in the original GP-UCB, we develop Safe-UCB without leveraging this information. A main difference between the original GP-UCB and Safe-UCB is that we add another GP to collect data of $g$ and learn the unsafe torque values by using the upper confidence bound as in ARTEO. The complete algorithm of Safe-UCB is given in \cref{sec:algsafeucb}. We simulate the reference current value in \cref{Figure6} with 50 different safe seeds. The results show that Safe-UCB tends to violate the safety constraint and operate the electric motors far from optimal values at first explored points due to high standard deviation in its predictions. Even though ARTEO is affected also from the same level of uncertainty at the beginning of the simulation, the standard deviation of its decisions is significantly less than Safe-UCB. 

\begin{figure}[tbhp]
  \centering
  \includegraphics[width=0.45\textwidth]{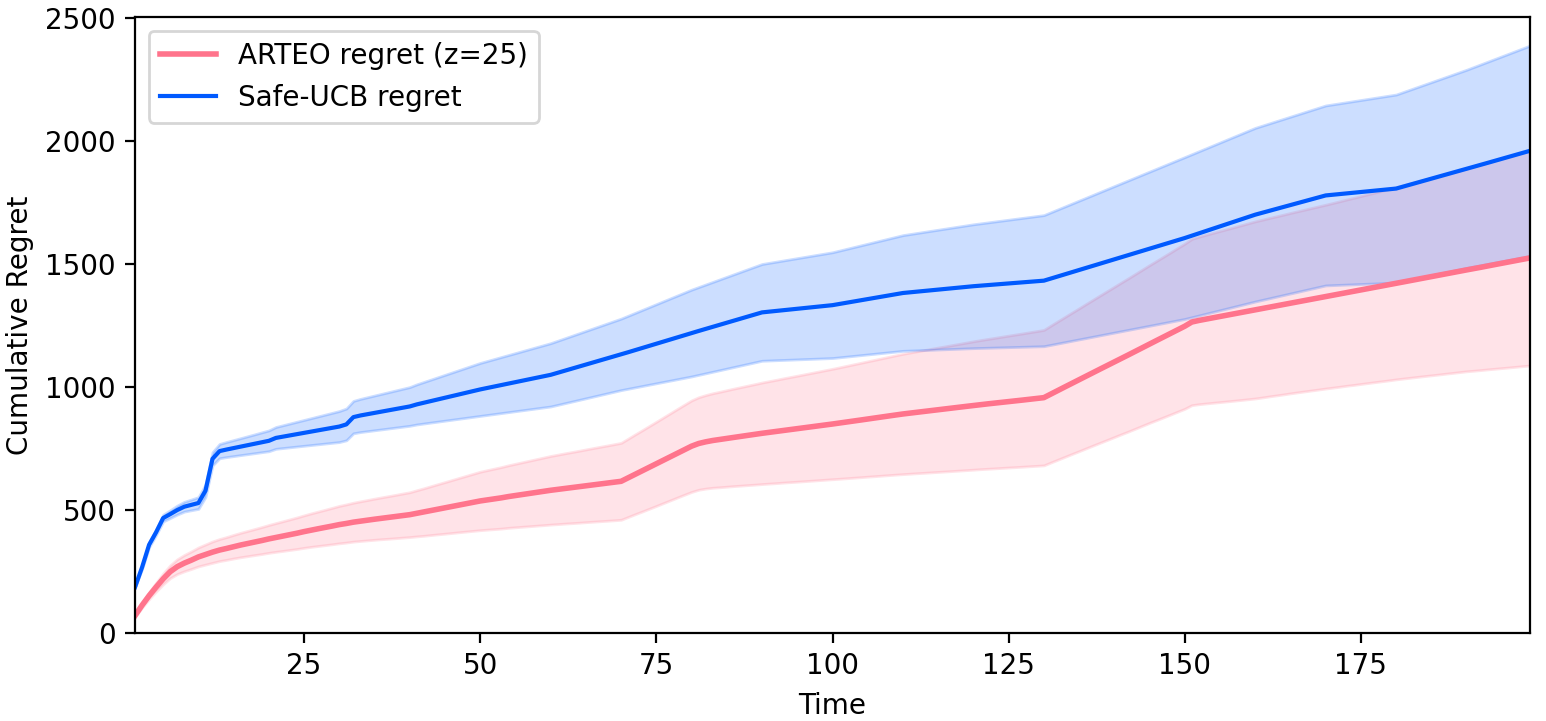}
  \caption{Cumulative regret of ARTEO and Safe-UCB with 50 different safe seeds. Shaded area represents $\pm0.1$ standard deviation.}
  \label{Figure9}
\end{figure}

Moreover, \cref{Figure6} demonstrates that while Safe-UCB under or over delivers produced current in several points, ARTEO is more capable to find points that minimize the objective function unless the reference value is unsafe with respect to safety constraint. We further compare the cumulative regret of both algorithms as in \cref{Figure9} by defining the regret $r_t$ at time step $t$, 
\begin{equation}\label{regretdef}
  r_t = |\max(Cr_t, 225.6) - \sum_{\lambda=1}^{2}\mu_{TC_\lambda}(X_t)|, \quad \forall t
\vspace{-0.5em}
\end{equation}
\cref{Figure9} shows the superiority of ARTEO to learn the unknown characteristics in a more accurate manner and leverage them in optimization.  

\subsection{Online Bid Optimization}

\begin{figure}[t]
  \centering
  \includegraphics[width=0.45\textwidth]{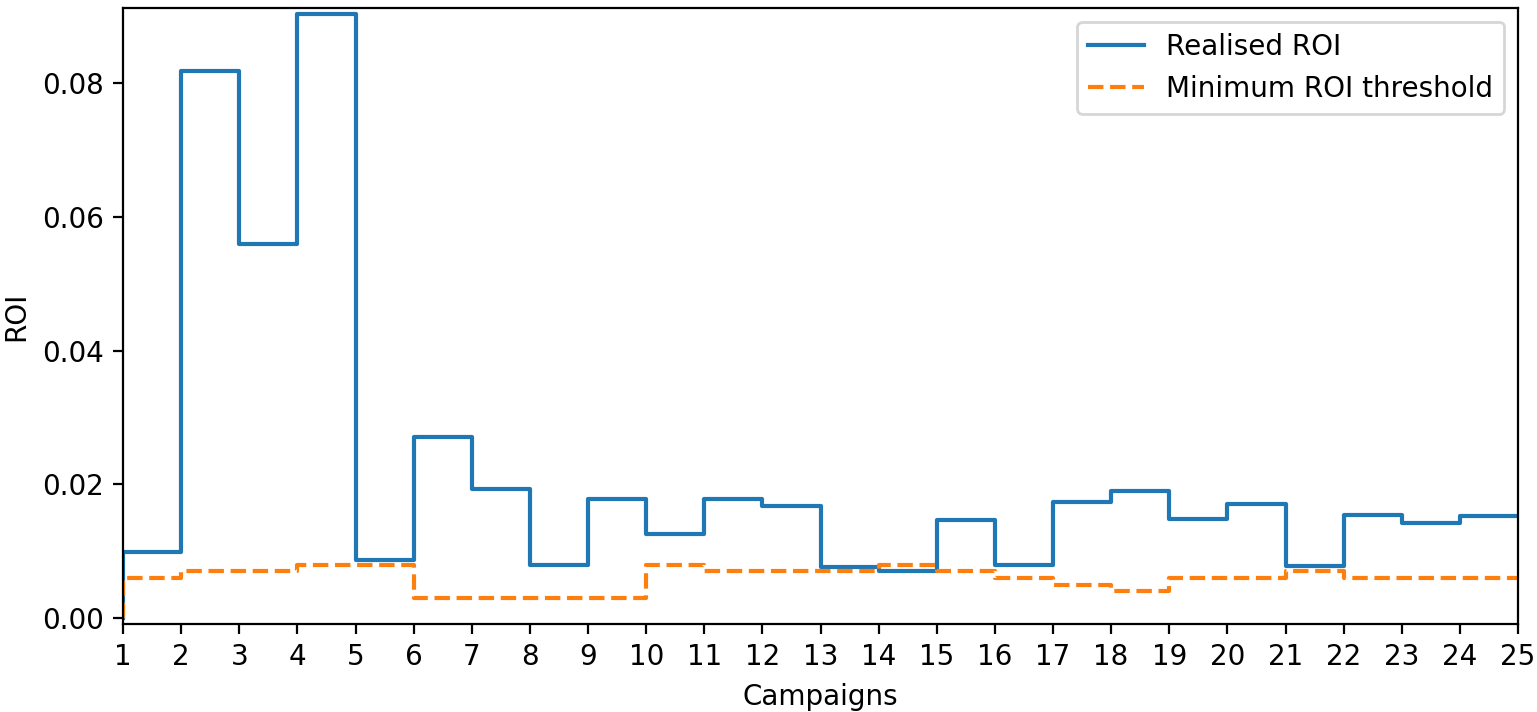}
  \caption{The minimum and achieved ROIs for campaigns. ARTEO is able to remain above the threshold for all campaigns.}
  \label{Figure7}
\end{figure}

In the second experiment, we investigate the implementation of ARTEO in a multi-dimensional problem of online bid optimization from the advertiser perspective. In bid optimization, the advertiser sets bid values with the aim of achieving high volumes by maximizing the number of shown advertisements and high profitability by maximizing the return-on-investment (ROI) ratio. In most agreements, unsatisfied ROI causes financial losses for advertisers. It becomes more challenging to sustain high ROIs when the number of advertisements increases. Constraining the ROI to remain above a certain threshold is a common approach, however, this method does not guarantee to satisfy the ROI constraint with zero violation \citep{SafeRTB}. ROI is measured by the revenues and costs, which are unknown to the bidding algorithms and brings uncertainty to the online bid optimization problem. Therefore, safe optimization algorithms could be useful to set bid values under the uncertainty of the revenues. 

We apply ARTEO to the iPinYou dataset \citep{ipinyoudataset}. This dataset has been released by a leading DSP (Demand-Side Platform) in China and consists of relevant information for personalized ads such as creative metadata, interests of users, and advertisement slot properties with decided bid price by their internal algorithm. It has been widely used as a benchmark to evaluate the performance of real-time-bidding algorithms \citep{ipinyoudatasetex3,ipinyoudatasetex2,ipinyoudatasetex1}. We simulate our approach by creating different campaign subsets from the original data, and for each campaign $t$, we minimize the following cost function
\begin{equation}\label{bidotobj}
\begin{split}
f{(X_t)} = \sum_{j=1}^{m}{cx_{tj}\mu_{C}(x_{tj})}  + \sum_{j=1}^{m}{\sqrt{(x_{tj} - \mu_{BP}(x_{tj}))^2}} - \\ z\sum_{j=1}^{m}{\bigl(\sigma_{C}(x_{tj})+\sigma_{BP}(x_{tj})\bigr)} \quad \quad
\end{split}
\raisetag{22pt}
\end{equation}
where $j$ denotes the ad number in the campaign, $\mu_{C}(x_{tj})$ is the mean prediction of GP for the $j$th advertisement to get a click with set bid values $x$, and $\mu_{BP}(x_{tj})$ is the mean prediction of GP for the bid price of $j$th ad in campaign $t$ based on previous observations. The fixed budget constraint for $m$ number of ads in campaign $t$ is formulated as
% \vspace{-1em}
\begin{equation}\label{fixedbudgetconst}
\begin{split}
&\sum_{j=1}^{m}{x_{tj}} \leq 180 m, \quad \forall t
\end{split}
\end{equation}
% \vspace{-0.5em}
The safe ROI constraint is constructed for the threshold $h_t$ for the campaign $t$ as follows
\begin{equation}\label{bidoptsafetyconst}
\begin{split}
&{\frac{\sum_{j=1}^{m}\mu_{C}(x_{tj}) - \beta_t \sigma_{C}(x_{tj})}{\sum_{j=1}^{m}{x_{tj}}}} \geq h_t 
\end{split}
\end{equation}
Lastly, the bid values $x_{tj}$ are bounded with non-negativity for all campaigns $t$ and for all advertisements $j$.

As opposed to the first experiment, where the changes in optimization goals were driven by changes in current references, an increase in $t$ in the online bid optimization example is driven by starting a new campaign. We construct two GPs in this experiment, the first GP learns the bid prices from past observations, and the second one models the impressions, which are represented in binary for clicks. The impressions are traditionally predicted by classifiers due to their binary representation. However, it is possible to cast it as a regression problem where we decide the binary representations after thresholding. Since we use covariance functions of GPs to model uncertainty, we cast it as a regression and guide our RTO with continuous values. The optimization algorithm bids an ad comparatively high when its value is higher than others. 

\begin{figure}[t]
  \centering
  \includegraphics[width=0.45\textwidth]{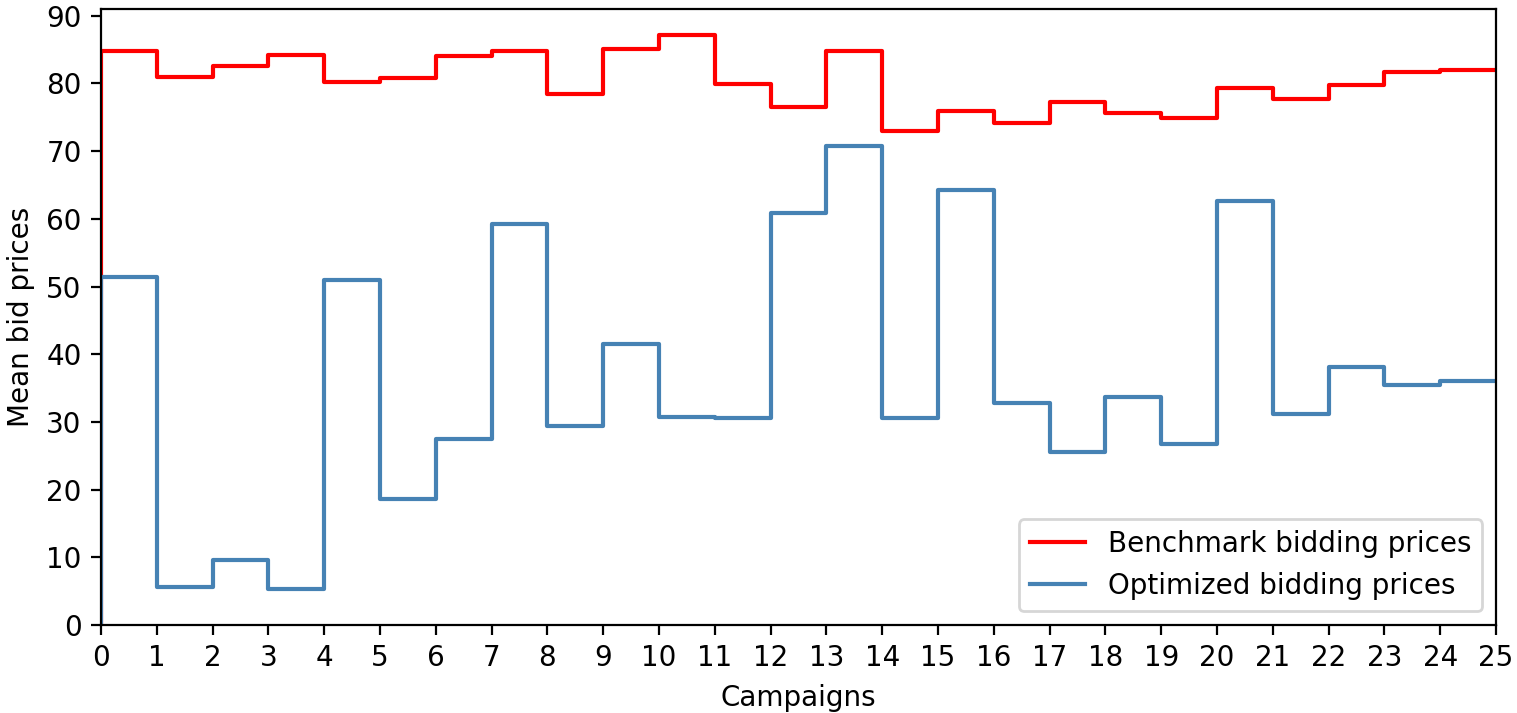}
  \caption{The mean bid prices for the given benchmark and ARTEO. ARTEO achieves higher ROI with lower costs.}
  \label{Figure8}
\end{figure}

Different feature sets in the dataset are used to compute posteriors based on relevance to the predictions. Further implementation details such as kernel function, hyperparameters and safety limits could be found in \cref{detailsRTB}. The results of the simulation are given in Figure \ref{Figure7} and Figure \ref{Figure8}. Our approach remains above the safety threshold while proposing lower bid prices compared to the given bid prices of the algorithm in \citet{ipinyoudataset}. 
\section{CONCLUSIONS}

In this work, we investigate real-time optimization of sequential-decision making in a safety-critical system under uncertainty due to partially known characteristics. We propose a safe exploration and optimization algorithm which follows the optimization goals and does exploration without violating the constraints. We adaptively control the contribution of exploration via the uncertainty term in the cost function. We demonstrate the applicability of our approach to real-world case studies. Our algorithm is able to model the unknown parameters in both applications and satisfy the optimization goals and safety constraints.

%%%%%%%%%%%%%%%%%%%%%%%%%%%%%%%%%%%%%%%%%%%%%%%%%%%%%%%%%%%%%%%%%%%%%%%%%%%%%%%%

%%%%%%%%%%%%%%%%%%%%%%%%%%%%%%%%%%%%%%%%%%%%%%%%%%%%%%%%%%%%%%%%%%%%%%%%%%%%%%%%

%%%%%%%%%%%%%%%%%%%%%%%%%%%%%%%%%%%%%%%%%%%%%%%%%%%%%%%%%%%%%%%%%%%%%%%%%%%%%%%%

%%%%%%%%%%%%%%%%%%%%%%%%%%%%%%%%%%%%%%%%%%%%%%%%%%%%%%%%%%%%%%%%%%%%%%%%%%%%%%%%

\bibliography{bibliographyfile}
\bibliographystyle{unsrtnat}

% \bibliographystyle{icml2023}

% %%%%%%%%%%%%%%%%%%%%%%%%%%%%%%%%%%%%%%%%%%%%%%%%%%%%%%%%%%%%%%%%%%%%%%%%%%%%%%%
%%%%%%%%%%%%%%%%%%%%%%%%%%%%%%%%%%%%%%%%%%%%%%%%%%%%%%%%%%%%%%%%%%%%%%%%%%%%%%%
% APPENDIX
%%%%%%%%%%%%%%%%%%%%%%%%%%%%%%%%%%%%%%%%%%%%%%%%%%%%%%%%%%%%%%%%%%%%%%%%%%%%%%%
%%%%%%%%%%%%%%%%%%%%%%%%%%%%%%%%%%%%%%%%%%%%%%%%%%%%%%%%%%%%%%%%%%%%%%%%%%%%%%%
\newpage
\appendix
\onecolumn
\section{Proofs}\label{appendixproofs}
\subsection{Proof of \cref{lem:lemma3}}
Let $\hat{p}^L = \mu(x_t) - \beta_t \sigma_{t}(x_t)$ and $\hat{p}^U = \mu(x_t) + \beta_t \sigma_{t}(x_t)$ where $\hat{p}^L \leq \hat{p}(x) \leq \hat{p}^U$. Given $g$ is monotonically related to $\hat{p}$, $g$ is in a known functional form of $g(\Delta(x), \hat{p}(x))$ with a known value of $\Delta(x)$, and $\hat{p}^L \leq \hat{p}(x) \leq \hat{p}^U$:
  \begin{itemize}
      \item if $g$ is monotonically related to $\hat{p}$ $\Rightarrow$ $g(\Delta(x), \hat{p}^L) \leq g(\Delta(x), \hat{p}(x)) \leq g(\Delta(x), \hat{p}^U)$.
      \item if $g$ is inversely monotonically related to $\hat{p}$ $\Rightarrow$ $g(\Delta(x), \hat{p}^U) \leq g(\Delta(x), \hat{p}(x)) \leq g(\Delta(x), \hat{p}^L)$.
  \end{itemize}
\begin{proof}\label{proof:lemma3}
    Let $g$, $\Delta$ and $p$ are continuous functions over domain $D$. We assume $g$ has a known functional form as defined by algebraic operations over known $\Delta(x)$ and unknown $\hat{p}(x)$, which is modelled by using Gaussian processes. It is given that $g$ is monotonically related to $p$. 
    \begin{enumerate}
    \item Consider the first case in \cref{lem:lemma3} as for any $x,y \in D$ such that $\hat{p}(x) \leq \hat{p}(y) \Rightarrow g(\cdot, \hat{p}(x))\leq g(\cdot, \hat{p}(y))$ (\cref{def:monotonicallyrelated}). For chosen $x_1, x_2, x_3 \in D$ such that
    \begin{equation}
        x_1 \leq x_2 \leq x_3 \Rightarrow \hat{p}(x_1) \leq \hat{p}(x_2) \leq \hat{p}(x_3).
    \end{equation}
    With given $g$ is monotonically related to $p$:
    \begin{equation}
    \label{firstcaseineq}
       \hat{p}(x_1) \leq \hat{p}(x_2) \leq \hat{p}(x_3) \Rightarrow g(\cdot, \hat{p}(x_1)) \leq g(\cdot, \hat{p}(x_2)) \leq g(\cdot, \hat{p}(x_3)). 
    \end{equation}

    Let $\hat{p}^L = \hat{p}(x_1)$ and $\hat{p}^U = \hat{p}(x_3)$. By substituting terms in  \cref{firstcaseineq}, we obtain
     \begin{equation}
       \hat{p}^L \leq \hat{p}(x_2) \leq \hat{p}^U \Rightarrow g(\cdot, \hat{p}^L) \leq g(\cdot, \hat{p}(x_2)) \leq g(\cdot, \hat{p}^U). 
    \end{equation}

    We can replace $x_2$ with any $x \in D$ that satisfies $\hat{p}^L \leq \hat{p}(x) \leq \hat{p}^U$. Thus, we have:

      \begin{equation}
       \hat{p}^L \leq \hat{p}(x) \leq \hat{p}^U \Rightarrow g(\cdot, \hat{p}^L) \leq g(\cdot, \hat{p}(x)) \leq g(\cdot, \hat{p}^U).
    \end{equation}

    \item Consider the second case in \cref{lem:lemma3} as for any $x,y \in D$ such that $\hat{p}(x) \leq \hat{p}(y) \Rightarrow g(\cdot, \hat{p}(x))\geq g(\cdot, \hat{p}(y))$ (\cref{def:inverselymonotonicallyrelated}). For chosen $x_1, x_2, x_3 \in D$ such that  
    \begin{equation}
        x_1 \leq x_2 \leq x_3 \Rightarrow \hat{p}(x_1) \leq \hat{p}(x_2) \leq \hat{p}(x_3).
    \end{equation}
    With given $g$ is inversely monotonically related to $p$:
    \begin{equation}\label{secondcaseineq}
       \hat{p}(x_1) \leq \hat{p}(x_2) \leq \hat{p}(x_3) \Rightarrow g(\cdot, \hat{p}(x_1)) \geq g(\cdot, \hat{p}(x_2)) \geq g(\cdot, \hat{p}(x_3)). 
    \end{equation}

    Let $\hat{p}^L = \hat{p}(x_1)$ and $\hat{p}^U = \hat{p}(x_3)$. By substituting terms in  \cref{secondcaseineq}, we obtain
     \begin{equation}
       \hat{p}^L \leq \hat{p}(x_2) \leq \hat{p}^U \Rightarrow g(\cdot, \hat{p}^U) \leq g(\cdot, \hat{p}(x_2)) \leq g(\cdot, \hat{p}^L). 
    \end{equation}

    We can replace $x_2$ with any $x \in D$ that satisfies $\hat{p}^L \leq \hat{p}(x) \leq \hat{p}^U$. Thus, we have:

      \begin{equation}
       \hat{p}^L \leq \hat{p}(x) \leq \hat{p}^U \Rightarrow g(\cdot, \hat{p}^U) \leq g(\cdot, \hat{p}(x)) \leq g(\cdot, p^L). 
    \end{equation}
    \end{enumerate}
\end{proof}

\subsection{Proof of \cref{thm:safetytheorem}}
Suppose that $\hat{p}$ and $g$ are continuous on compact set $D$, the functional form of $g(\Delta(x), \hat{p}(x))$ is known and $g$ is monotonically related to $\hat{p}$ where $\hat{p}$ is modelled from a GP through noisy observations $y_t = \hat{p}(x_t) + \epsilon_t $ and $\epsilon_t$ is a $R$-sub-Gaussian noise for a constant $R \geq 0$ at each iteration $t$. For a known value of $\Delta(x)$, the maximum and minimum values of $g(\Delta(x), \hat{p}(x))$ lie on the upper and lower confidence bounds of the Gaussian process obtained for $\hat{p}(x)$ which are computed as in \cref{lem:lemma3}. For a chosen $\beta_t$ and allowed failure probability $\delta$ as in \cref{betateq}, 
\begin{itemize}
    \item if $g$ is monotonically related to $\hat{p} \Rightarrow P\bigr[ g(\Delta(x_t), \hat{p}^L) \leq g (\Delta(x_t), \hat{p}(x_t)) \leq g(\Delta(x_t), \hat{p}^U) \bigr] \leq 1 - \delta, \forall t \geq 1$
    \item if $g$ is inversely monotonically related to $\hat{p} \Rightarrow P\bigr[ g(\Delta(x_t), \hat{p}^U) \leq g (\Delta(x_t), \hat{p}(x_t)) \leq g(\Delta(x_t), \hat{p}^L) \bigr] \leq 1 - \delta, \forall t \geq 1$
\end{itemize}

\begin{proof}
    Theorem 2 by \citet{BoundProof} shows that the following holds with probability at least $1 - \delta$:
    \begin{equation}
        \forall t \geq 1 \ \forall x \in D, |\hat{p}(x) - \mu_{t-1}(x)| \leq \beta_t\sigma_{t-1}(x),
    \end{equation}
     by choosing a $\beta_t$ as in \cref{betateq} under the assumptions of $\|p\|_k \leq B$ and $\epsilon_t$ is $R$-sub-Gaussian for all $t \geq 1$ (for proof, see Theorem 2 of \citep{BoundProof}). We can extract the inequality from absolute value as in the following
     \begin{equation}
     \label{maininequality} 
          \mu_{t-1}(x)-\beta_t\sigma_{t-1}(x) \leq \hat{p}(x) \leq \mu_{t-1}(x)+\beta_t\sigma_{t-1}(x).
     \end{equation}  
     Define $\hat{p}^L$ and $\hat{p}^U$ as
     \begin{equation}
          \hat{p}^L=  \mu_{t-1}(x)-\beta_t\sigma_{t-1}(x).
     \end{equation}
     \begin{equation}
          \hat{p}^U =  \mu_{t-1}(x)+\beta_t\sigma_{t-1}(x).
     \end{equation}\
    Then, plug $\hat{p}^L$ and $\hat{p}^U$ into \cref{maininequality} and obtain the following with $1 - \delta$ probability
    \begin{equation}
        \hat{p}^L \leq \hat{p}(x) \leq \hat{p}^U.
    \end{equation}
    By the proof of \cref{lem:lemma3}, we can reflect this inequality to $g$ since $g$ is defined as monotonically related to $p$. Thus following statements hold with $1 - \delta$ probability,
    \begin{itemize}
       \item if $g$ is monotonically related to $p$ $\Rightarrow g(\Delta(x_t), \hat{p}^L) \leq g (\Delta(x_t), \hat{p}(x_t)) \leq g(\Delta(x_t), \hat{p}^U) $,
        \item if $g$ is inversely monotonically related to $p$  $\Rightarrow g(\Delta(x_t), \hat{p}^U) \leq g (\Delta(x_t), \hat{p}(x_t)) \leq g(\Delta(x_t), \hat{p}^L)$.
    \end{itemize}
    
\end{proof}

\section{Experiment Details}

We have constrained nonlinear problems in the experiments section, and we choose interior-point and sequential-least square programming (SLSQP) algorithms to solve our first and second problems, respectively. 

\subsection{Implementation of ARTEO to Electric Motor Current Optimization}\label{detailsCurrentOpt}

\subsubsection{Electric Motor Simulation Environment Details}\label{gem-details}

Gym-Electric-Motor (GEM) is an environment that includes the simulation of different types of electric motors with adjustable parameters such as load, speed, current, torque, etc. to train reinforcement learning agents or build model predictive control solutions to control the current, torque or speed for a given reference. In electric motors, the operation range is limited by nominal values of variables to prevent motor damage. Furthermore, there are also safety limits for some parameters, such as the maximum safe current limit to avoid excessive heat generation.

\vskip -0.15in
\begin{table}[h]
\caption{Electric motor parameters}
\label{parameters-table}
% \vskip 0.15in
\begin{center}
\begin{small}
\begin{sc}
\begin{tabular}{lccccr}
\toprule
Electric Motor & $R_a$ & $L_a$ & $\psi_e$ & $J_{rotor}$\\
\midrule
Machine-1 & 0.016 & 1.9e-05 & 0.165 & 0.025 \\
Machine-2 & 0.01 & 1.5e-05 & 0.165 & 0.025 \\
\bottomrule
\end{tabular}
\end{sc}
\end{small}
\end{center}
\vskip -0.1in
\end{table}

\subsubsection{The Impact of Exploration Hyperparameter}

\begin{figure}[tbhp]
  \centering
  \includegraphics[width=0.45\textwidth]{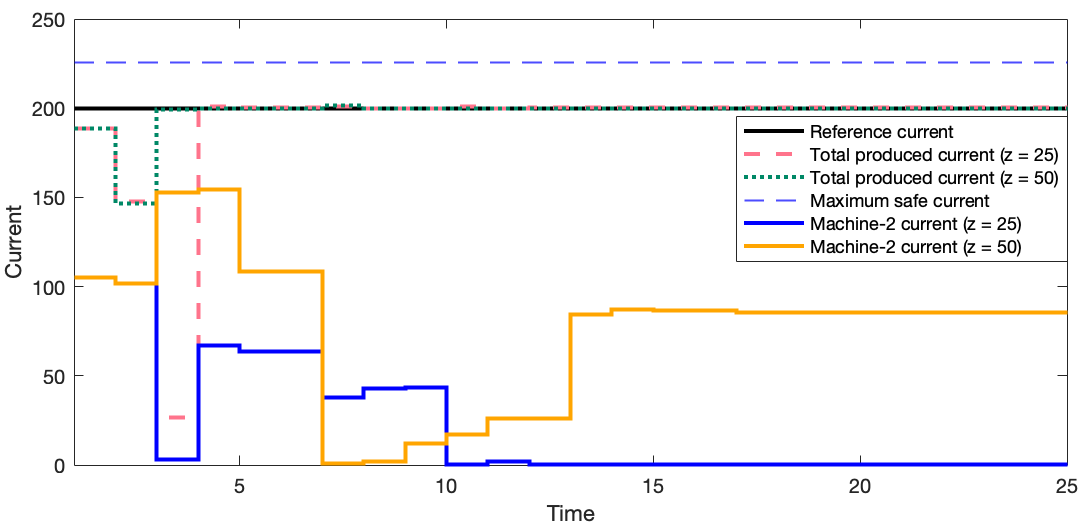}
  \caption{Greater $z$ values encourage exploration at points with a greater standard deviation.}
  \label{Figure4}
\end{figure}

\begin{figure}[tbhp]
  \centering
  \includegraphics[width=0.45\textwidth]{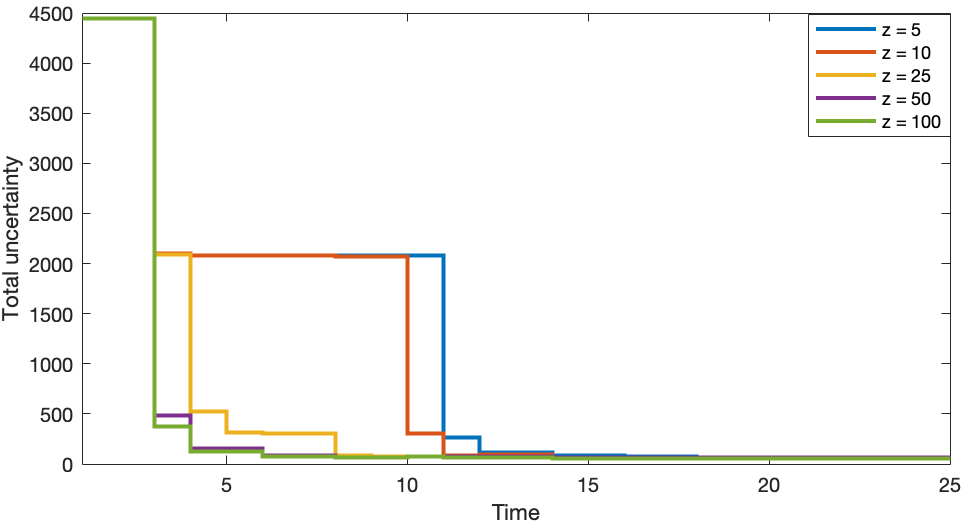}
  \caption{Total uncertainty decreases with a greater rate over time for greater $z$ values in an environment with a constant reference.}
  \label{Figure5}
\end{figure}

The exploration in ARTEO is driven by the $z$ hyperparameter. It is possible to create different strategies according to the requirements of the problem by setting different values to this hyperparameter. To demonstrate this, we simulate a constant reference current scenario with different $z$ values, as in \cref{Figure4}. This figure exhibits that greater $z$ values lead to more frequent changes in reference torque values while preserving the ability to satisfy the reference current. It is expected that more frequent changes in the operating points assist in decreasing the total uncertainty in the environment faster. This is demonstrated in \cref{Figure5} for different $z$ values for the reference scenario in \cref{Figure4}. Hyperparameter optimization methods could be leveraged to choose the value of $z$ to achieve minimum regret where regret at time $t$ is defined as \cref{regretdef}.

\subsubsection{Hyperparameter Optimization for $z$}\label{hyperparameteropt}

\begin{figure}[tbhp]
  \centering
  \includegraphics[width=0.43\textwidth]{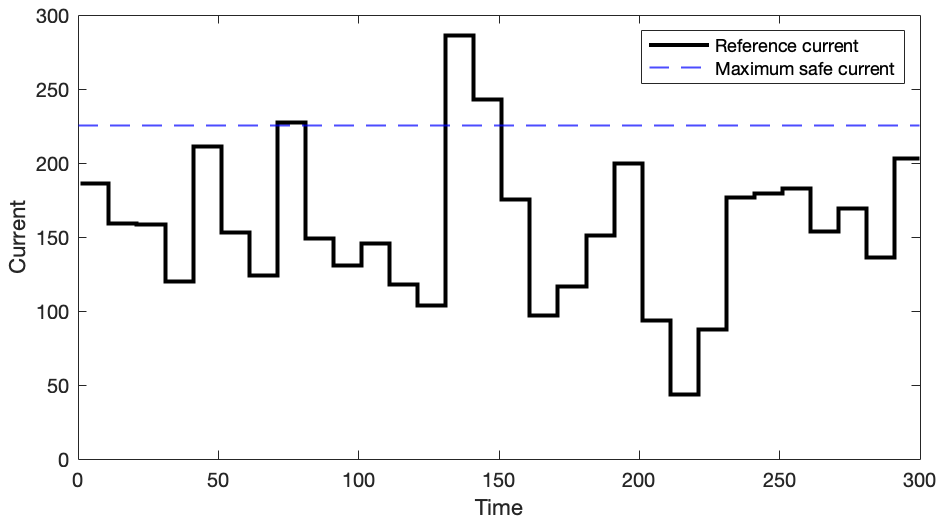}
  \caption{Reference current of the hyperparameter optimization simulation (longer simulation of \cref{Figure6}).}
  \label{Figure10}
\end{figure}

\begin{figure}[tbhp]
  \centering
  \includegraphics[width=0.48\textwidth]{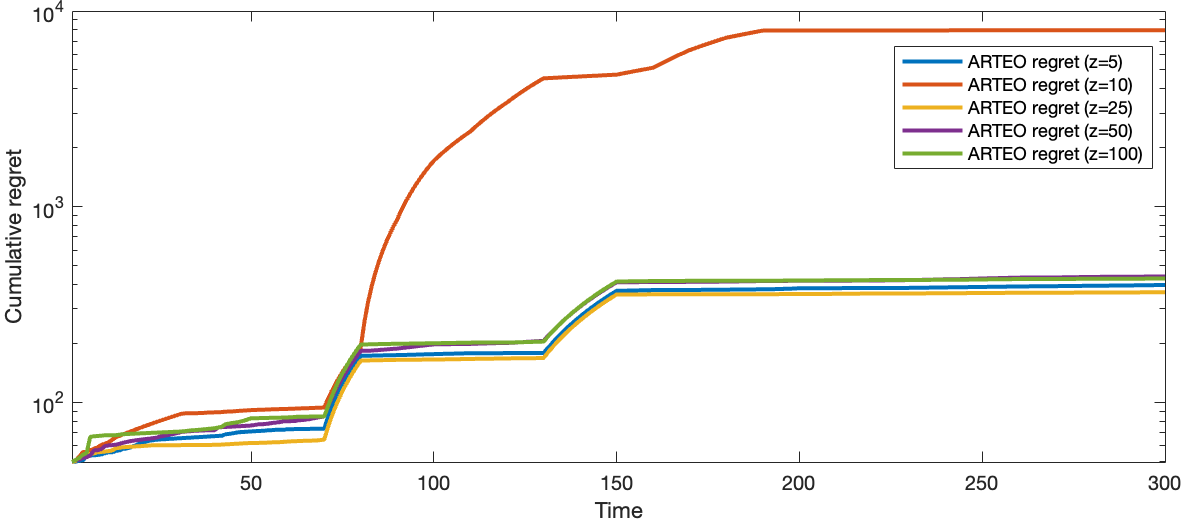}
  \includegraphics[width=0.43\textwidth]{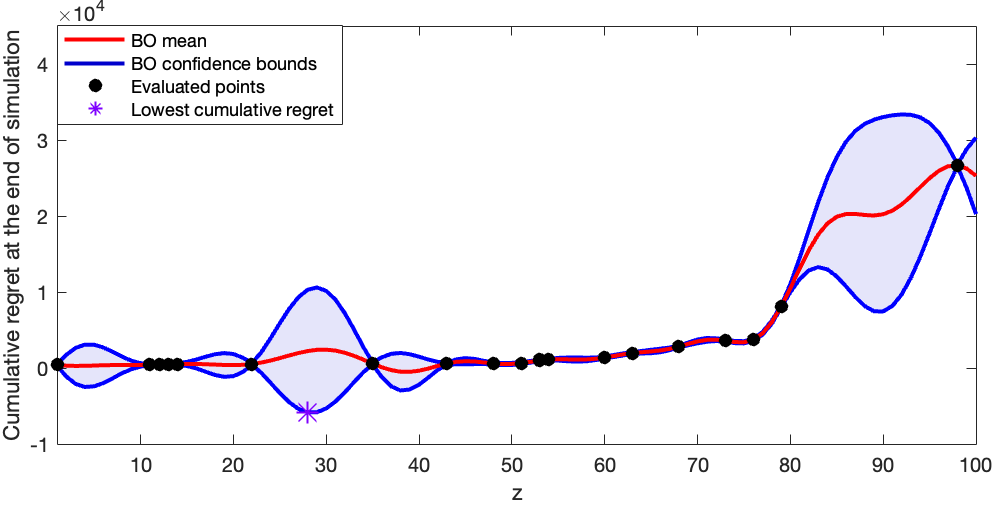}
  \caption{Hyperparameter optimization results for $z$ in electric motor current optimization with reference scenario in \cref{Figure10}. The left figure depicts the cumulative regret (in log scale) over time for each evaluated $z$ value in grid-search. The right figure shows the evaluated $z$ values in BO with lower confidence bound acquisition function and fit GP model mean and confidence bounds. }
  \label{Figure11}
\end{figure}

During the implementation of ARTEO to electric motor current optimization, we choose $z = 25$ after applying two hyperparameter optimization methods as the grid-search \citep{GridSearch} and Bayesian optimization (BO) \citep{BO}. We evaluate different $z$ values based on the cumulative regret at the end of the simulation of reference in \cref{Figure10}. For grid-search, we evaluate the cumulative regret with $z$ taking values of 5, 10, 25, 50, and 100. The results in \cref{Figure11} show that the most suitable $z$ value for the given reference is 25 amongst the evaluated values. 

As an alternative hyperparameter technique to grid-search, BO is also applied to the simulation of the given reference. BO is an optimization method that builds a surrogate model with evaluations at chosen points and then chooses the next value to be evaluated based on the minimization/maximization of the chosen acquisition function, which is specified as lower confidence bound in our implementation. Further details of BO could be found in \citet{BO}. We limit the maximum number of evaluations with 35 points and the results in \cref{Figure11} demonstrate that the surrogate model of BO suggests that the minimum cumulative regret at the end of the simulation (of \cref{Figure10}) when $z=28$. 

Hyperparameter optimization becomes more challenging in the online learning setup where complete information is not available to do simulation. A suggested hyperparameter optimization method for ARTEO implementation in online learning could be starting with a random value and adjusting it based on replaying the available simulation and optimizing the hyperparameter $z$ asynchronously whenever new information is available. Similar hyperparameter optimization techniques for online learning could be found in \citet{AsyncHO} and \citet{HyperparameterOptimizationOnline}.

\subsection{Implementation of Safe-UCB to Electric Motor Current Optimization}\label{sec:algsafeucb}

In electric motor current optimization experiments, $f$ is calculated as the difference between the given reference current and the total produced current. Thus, we search points that minimize the value of $f$ which leads us to use lower confidence bound of $f$ by following the optimism principle of \citet{1985Lai}. The safety function $g$ is defined as the difference between the safety limit value $h = 225.6$ and the total produced current. Hence, the chosen points are safe as long as the $g$ value of chosen points remains above zero. 

% \begin{figure}[t]
\begin{algorithm}[H]
  \caption{Safe-UCB}\label{alg:Safe-UCB}
  \begin{algorithmic}
  \STATE {\bfseries Input:} Decision set $D$ for each variable $i \in \left\{1,..,n\right\}$, GP priors for ${GP}^{f}$ and ${GP}^{g}$, safe seed sets $S^f_{0}$ and $S^g_{0}$ 
  \FOR{$t=1,...,T$} 
  \STATE Update ${GP}^{f}$ by conditioning on $S_{t-1}^f$
  \STATE Update ${GP}^{g}$ by conditioning on $S_{t-1}^g$ 
  \STATE Choose $x_{t}^{*} = \argmin_{x \in D} \mu^{f}_{t}(x) - \beta_t \sigma^{f}_t(x)$ subject to $\mu^{g}_{t}(x) - \beta_t \sigma^{g}_t(x) \geq 0$
  \STATE $y_{t}^f \gets$ $f(x_{t}^*) + \epsilon_{t}^f$
  \STATE $y_{t}^g \gets$ $g(x_{t}^*) + \epsilon_{t}^g$
  \STATE $S_{t}^f \gets$ $S^f_{t-1} \cup \left\{x^*_{t}:{y^f_{t}}\right\} $ 
  \STATE $S_{t}^g \gets$ $S^g_{t-1} \cup \left\{x^*_{t}:{y^g_{t}}\right\} $ 
  \ENDFOR
  \end{algorithmic}
  \end{algorithm}
  % \end{figure}

\subsection{Environment Details for Online Bid Optimization}\label{detailsRTB}

The GP of the bid price is initialized with the Matern kernel with $\nu$ = 1.5 and is trained over 143 features whereas the impression GP has the Squared Exponential kernel and 69 features. The safe seeds start with 30 samples, which is higher than our first experiment since this is a higher dimensional problem. As a minimum ROI threshold, 90\% of the given benchmark data ROI is set due to having a strict budget and ROI requirements in our setup. We partition the selected subset of the dataset into 25 campaigns. Each campaign has its ROI threshold and budget, which are calculated as \cref{fixedbudgetconst} and \cref{bidoptsafetyconst}.

The algorithm starts with a safe seed set to compute the posteriors of GPs, and then for each campaign, it utilizes the mean and standard deviation of GP posteriors to measure ROI and click probability. During the RTO phase, the higher bid prices for higher estimated click values are encouraged within a fixed budget, and the difference between the predicted bid price by GP and the proposed bid price by RTO for each advertisement is accumulated and introduced as a penalty in the objective function. Thus, the algorithm does not put the entire budget into the highest-valued ad within the campaign. At the end of each campaign, true bid prices and clicks with additive Gaussian noise are used to update the posteriors of GPs. The feedback is given only for ads in the campaign with a non-negative optimized bid price which leads to high standard deviations for non-bid similar ads. The environment becomes available for exploration after spending less than the sum of predicted bid prices and satisfying minimum thresholds in two consecutive campaigns. Hence, $z$ is set to $\zeta$ = 100 to excite the RTO to take decisions at points that could reduce uncertainty in predictions.

\end{document}